\documentclass[letterpaper, 10 pt, journal, twoside]{ieeetran}
\IEEEoverridecommandlockouts    

\usepackage{hyperref}
\usepackage{makecell}
\usepackage{threeparttable}
\usepackage{caption}
\usepackage{soul}
\usepackage{color}
\usepackage[table]{xcolor}
\usepackage{colortbl}
\hypersetup{
    colorlinks=true,
    linkcolor=magenta,
    filecolor=magenta,      
    urlcolor=magenta,
}

\usepackage{graphics}           
\usepackage{times}              
\usepackage{amsmath}            
\usepackage{amssymb}            
\usepackage{graphicx}
\usepackage{algorithm}
\usepackage[noend]{algpseudocode}
\usepackage{booktabs}
\usepackage{color}
\definecolor{instructioncolor}{rgb}{.5,.5,.5}

\usepackage[font=small]{caption}

\def\figref#1{Fig.~\ref{#1}}
\def\tabref#1{Tab.~\ref{#1}}
\def\eqref#1{Eq.~(\ref{#1})}

\makeatletter
\usepackage{xspace}
\DeclareRobustCommand\onedot{\futurelet\@let@token\@onedot}
\def\@onedot{\ifx\@let@token.\else.\null\fi\xspace}

\makeatother

\usepackage{array}
\newcolumntype{L}[1]{>{\raggedright\let\newline\\\arraybackslash\hspace{0pt}}m{#1}}
\newcolumntype{C}[1]{>{\centering\let\newline\\\arraybackslash\hspace{0pt}}m{#1}}
\newcolumntype{R}[1]{>{\raggedleft\let\newline\\\arraybackslash\hspace{0pt}}m{#1}}

\usepackage{bm}
\usepackage{multirow}
\usepackage{rotating}  

\title{LRFusionPR: A Polar BEV-Based LiDAR-Radar Fusion Network for Place Recognition}

\author{Zhangshuo Qi$^{1}$, Luqi Cheng$^{1}$, Zijie Zhou$^{1}$, and Guangming Xiong$^{*}$
\thanks{Manuscript received: April, 26, 2025; Revised July, 28, 2025; Accepted September, 14, 2025.}
\thanks{This paper was recommended for publication by Editor Pascal Vasseur upon evaluation of the Associate Editor and Reviewers' comments.
This work was supported by the National Natural Science Foundation of China under Grant 52372404.} 
\thanks{$^{1}$Zhangshuo Qi, Luqi Cheng, and Zijie Zhou are with  Beijing Institute of Technology, Beijing, 100081, China
        {\tt\footnotesize 3120240323@bit.edu.cn}}%
\thanks{$^{*}$Guangming Xiong (corresponding author) is with Beijing Institute of Technology, Beijing, 100081, China
        {\tt\footnotesize xiongguangming@bit.edu.cn}}%
\thanks{Digital Object Identifier (DOI): see top of this page.}
}

\begin{document}
\maketitle

\IEEEpeerreviewmaketitle

\begin{abstract}

In autonomous driving, place recognition is critical for global localization in GPS-denied environments. LiDAR and radar-based place recognition methods have garnered increasing attention, as LiDAR provides precise ranging, whereas radar excels in adverse weather resilience. However, effectively leveraging LiDAR-radar fusion for place recognition remains challenging. The noisy and sparse nature of radar data limits its potential to further improve recognition accuracy. In addition, heterogeneous radar configurations complicate the development of unified cross-modality fusion frameworks. In this paper, we propose LRFusionPR, which improves recognition accuracy and robustness by fusing LiDAR with either single-chip or scanning radar. Technically, a dual-branch network is proposed to fuse different modalities within the unified polar coordinate bird's eye view (BEV) representation. In the fusion branch, cross-attention is utilized to perform cross-modality feature interactions. The knowledge from the fusion branch is simultaneously transferred to the distillation branch, which takes radar as its only input to further improve the robustness. Ultimately, the descriptors from both branches are concatenated, producing the multimodal global descriptor for place retrieval. Extensive evaluations on multiple datasets demonstrate that our LRFusionPR achieves accurate and rotation-invariant place recognition, while maintaining robustness under varying weather conditions. Our open-source code is available at \url{https://github.com/QiZS-BIT/LRFusionPR}.\\
\end{abstract}

\section{Introduction}
\label{sec:intro}

\IEEEPARstart{P}{lace} recognition is a localization technique that uses perceptual data to identify whether a vehicle is revisiting a location~\cite{arandjelovic2016netvlad, uy2018pointnetvlad, suaftescu2020kidnapped, cait2022autoplace}. 
This capability enables global localization in GPS-denied environments and supports loop closure detection in SLAM systems. In autonomous driving applications, LiDAR place recognition (LPR) has attracted increasing attention~\cite{uy2018pointnetvlad, komorowski2021egonn, ma2022overlaptransformer, luo2024bevplace++}. LiDAR captures precise structural information and exhibits robustness under varying illumination conditions. These attributes endow LPR with significant potential.

Recent studies have focused on integrating LiDAR with other sensors, leveraging their complementary strengths to enhance recognition accuracy. The most prevalent multimodal place recognition (MPR) methods combine LiDAR with cameras~\cite{komorowski2021minkloc++,zhou2023lcpr}, which provide rich semantic and textural information. However, LiDAR and camera share a common drawback, exhibiting suboptimal performance under adverse weather conditions. 



\begin{figure}
  \centering
  \includegraphics[width=1\linewidth]{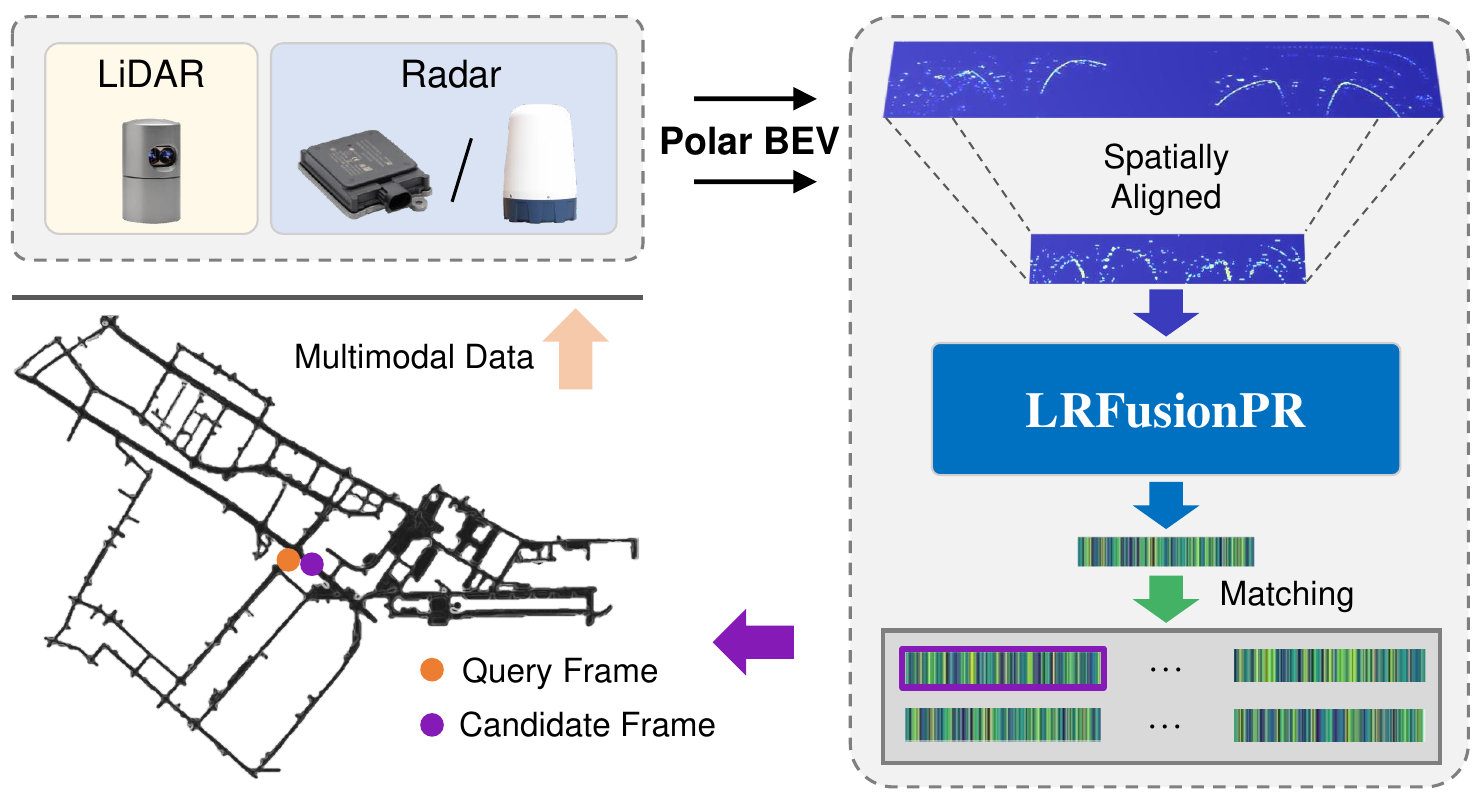}
  \caption{The fusion of LiDAR and radar enables the acquisition of fine-grained scene structure while improving weather robustness. In LRFusionPR, multimodal data are aligned into a polar BEV format, enabling feature fusion via cross-attention and cross-modality distillation. This results in multimodal global descriptors for accurate and robust place recognition.}
  \label{fig:motivation}
  \vspace{-0.65cm}
\end{figure}

To address these limitations, radar stands out as a promising alternative sensor. Although radar offers relatively lower data quality, its exceptional signal penetration capability provides superior weather robustness. However, how to effectively perform LiDAR-radar fusion for place recognition remains an open problem. The main challenges arise from two aspects. Firstly, the two most common types of radar in radar place recognition (RPR) differ in their perceptual formats. Single-chip radar generates sparse point clouds with velocity, while scanning radar outputs polar intensity maps. This domain gap limits existing RPR methods to specific radar types~\cite{suaftescu2020kidnapped, kim2024referee, cait2022autoplace}. Secondly, radar data exhibits sparse and noisy characteristics, which complicates the extraction of discriminative representations. 



In this paper, we propose LRFusionPR, a LiDAR-radar fusion-based MPR method designed to address these challenges, as shown in{~\figref{fig:motivation}}. The proposed framework demonstrates compatibility with heterogeneous radars, leveraging the individual strengths of both LiDAR and radar to deliver state-of-the-art recognition performance under adverse conditions. To address the challenge arising from heterogeneous data formats, we unify LiDAR and radar data formats through polar orthogonal BEV projection. Moreover, polar BEV facilitates the straightforward and effective extraction of rotation-invariant descriptors, a contrast to Cartesian BEV. Subsequently, a dual-branch network exploits cross-modality feature correlations within polar BEV representations. The fusion branch enhances recognition accuracy through cross-attention, integrating spatially aligned LiDAR and radar features. The distillation branch operates on radar-only inputs, fully unleashing the radar-specific robustness. Furthermore, to facilitate the extraction of discriminative radar features, we distill the knowledge from fusion branch into distillation branch. Ultimately, descriptors from both branches are concatenated for robust and accurate place recognition. 

In summary, our main contributions are as follows:
\begin{itemize}
\item We propose a polar BEV-based LiDAR-radar fusion network, LRFusionPR, to achieve accurate and robust place recognition in a rotation-invariant way.
\item We propose the LiDAR-radar feature fusion mechanism based on a dual-branch network. Cross-attention and distillation are synergistically integrated to preserve the specific advantages of both LiDAR and radar.
\item Extensive evaluations on four datasets and adverse weather conditions demonstrate our method's performance in both accuracy and robustness.
\end{itemize}

\section{Related Work}
\label{sec:related}

\subsection{LiDAR Place Recognition}
\label{sec:lpr}
LiDAR's robustness to illumination and seasonal variations makes LPR suitable for autonomous driving scenarios. Scan Context~\cite{kim2018scan} constructs descriptors by encoding the distribution of LiDAR point clouds. PointNetVLAD~\cite{uy2018pointnetvlad} takes downsampled point cloud submaps as input to perform point-wise feature extraction. OverlapTransformer~\cite{ma2022overlaptransformer} project point clouds into range images, enabling efficient feature extraction. MinkLoc3Dv2~\cite{komorowski2022improving} and EgoNN{~\cite{komorowski2021egonn}} utilizes 3D sparse convolution to extract spatial features from sparsely distributed LiDAR point clouds. BEVPlace++~\cite{luo2024bevplace++} project point clouds into BEV images to extract discriminative descriptors. More recently, the integration of LiDAR and camera has emerged as a paradigm for multimodal place recognition (MPR). MinkLoc++~\cite{komorowski2021minkloc++} enhance recognition accuracy by fusing descriptors from different modalities. LCPR~\cite{zhou2023lcpr} exploit cross-modality correspondence through interactive feature fusion. 

\subsection{Radar Place Recognition}
\label{sec:rpr}
While the radar offers relatively lower data quality than LiDAR, its robustness in adverse weather conditions has made RPR a viable solution. Single-chip radar and scanning radar are the two main radar types that have been extensively studied in RPR. Due to its comparatively dense and accurate nature, scanning radars have attracted considerable attention. KidnappedRadar~\cite{suaftescu2020kidnapped} achieves robust place recognition by extracting rotation-invariant features. Off the Radar{~\cite{yuan2023off}} introduces an uncertainty estimation-based approach for place recognition. Radar SLAM{~\cite{hong2021radar}} establishes a robust SLAM system capable of operating under all weather conditions. RaPlace{~\cite{jang2023raplace}} proposes a non-learning method utilizing the Radon transform. ReFeree{~\cite{kim2024referee}} extracts lightweight descriptors using a feature and free space.  OpenRadVLAD{~\cite{gadd2024open}} extracts discriminative descriptors from polar radar scans. Compared to scanning radar, single-chip radar data is sparser and noisier. AutoPlace~\cite{cait2022autoplace} improves the recognition accuracy by exploiting sequential information. CRPlace~\cite{fu2024crplace} proposes an MPR approach that integrates single-chip radar with camera. More recently, 4D RadarPR{~\cite{chen20254d}} proposed an RPR framework using 4D radar data.

\begin{figure*}[ht]
  \centering
  \includegraphics[width=1\linewidth]{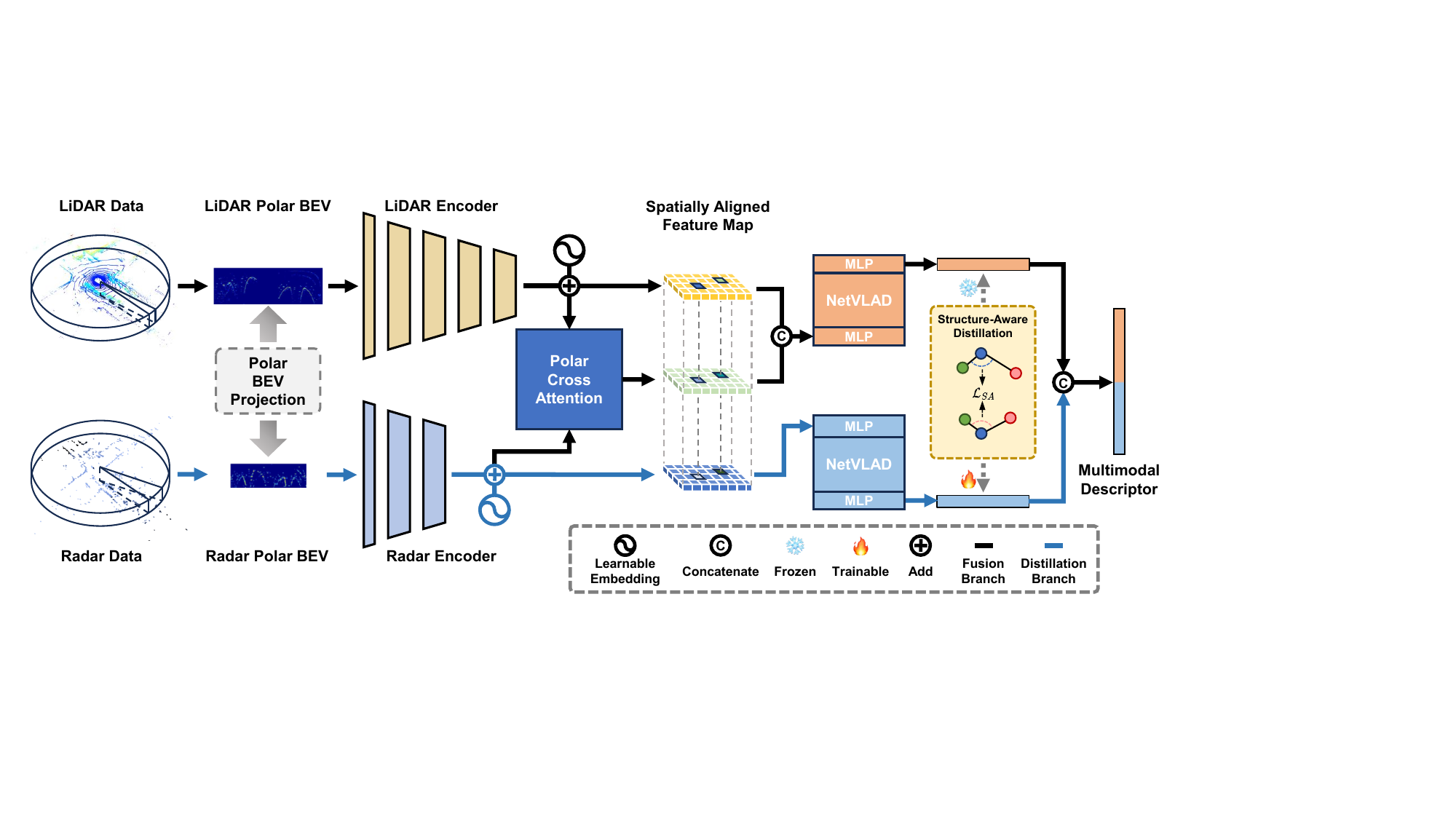}
  \caption{The overall architecture of LRFusionPR. The LiDAR and radar data are unified into polar BEV representations, and then fed into the ResNet encoders. Subsequently, the fusion branch exploits cross-modality feature correlations, and the distillation branch generates radar-based descriptors. The discriminability of the radar-based descriptors is enhanced by structure-aware distillation. Ultimately, the descriptors from both branches are concatenated to achieve accurate and robust place recognition.}
  \label{fig:overall}
  \vspace{-0.65cm}
\end{figure*}

\subsection{LiDAR-Radar Fusion}
\label{sec:lrfusion}
Recently, the combination of LiDAR and radar has attracted interest. LiDAR and cameras are prone to degradation in adverse weather, resulting in poor robustness. In contrast, radar’s weather resilience and LiDAR's high resolution form a strong complement. Extensive studies in 3D object detection have validated the advantages of LiDAR and radar fusion~\cite{qian2021robust}. In place recognition, some work, such as Radar-to-Lidar{~\cite{yin2021radar}}, has attempted to leverage the complementary strengths of LiDAR and radar through cross-modal alignment.

However, how to boost place recognition performance through LiDAR-radar fusion has not been fully explored. To address this, we propose a novel polar BEV-based MPR method, dubbed LRFusionPR. We aim to propose a framework capable of accommodating both single-chip and scanning radar, enabling accurate and robust place recognition under various weather conditions.

\section{Our Approach}
\label{sec:method}

\subsection{LRFusionPR Architecture}
\label{sec:architecture}
\figref{fig:overall} illustrates the architecture of our proposed LRFusionPR, which fuses LiDAR with single-chip or scanning radar for accurate and robust place retrieval. We first unify multimodal data into the polar BEV representation. Not only does it standardize the data formats between LiDAR and heterogeneous radars, it also provides a data foundation for rotation-invariant place recognition. Using two ResNet backbones~\cite{he2016deep} of different depths, we then extract spatially aligned features from polar BEVs. Subsequently, the multimodal features are fed into a dual-branch network, consisting of the fusion branch and the distillation branch. The main role of the fusion branch is to fully exploit the cross-modality feature correlations. Due to the inherent sparsity and noise of radar data, direct integration of radar features into a LiDAR-centric network may not be optimal. Therefore, we propose a polar cross-attention module to perform deep feature interaction between modalities. The discriminability of radar features is improved under the guidance of LiDAR cues. However, performing feature fusion alone cannot fully exploit the radar-specific robustness. To address this, we introduce the distillation branch, which processes radar data exclusively to generate robust descriptors. We further propose using the structure-aware loss~\cite{knights2022incloud} to guide knowledge transfer from the fusion branch to the distillation branch. This mechanism distills the multimodal fusion knowledge to the distillation branch, thereby increasing the discriminability of radar-based descriptors. Ultimately, the descriptors from both branches are concatenated to achieve accurate and robust place recognition in a rotation-invariant way.

\subsection{Polar BEV-Based Paradigm}
\label{sec:polar_bev}

\textbf{Polar orthogonal projection.} We unify LiDAR, single-chip radar, and scanning radar into a unified polar BEV format to address their inherent domain discrepancies.

Let LiDAR point cloud $\mathcal{P}$ and radar point cloud $\mathcal{R}$ constitute a spatially and temporally aligned input pair. By performing polar orthogonal projection on $\mathcal{P}$ and $\mathcal{R}$, the LiDAR polar BEV $\mathcal{B}_\text{P}$ and the radar polar BEV $\mathcal{B}_\text{R}$ are obtained respectively. Using the LiDAR modality as an example, each LiDAR point in the Cartesian coordinate $p_{i}=\left(x_{\text{P}}, y_{\text{P}}, z_{\text{P}}\right)^{\text{T}}$ is converted to the pixel coordinate $\left(u_{\text{P}}, v_{\text{P}}\right)$ of $\mathcal{B}_\text{P}$ by:
\begin{align}
    \left( \begin{array}{c} 
    u_{\text{P}} \\ 
    v_{\text{P}} 
    \end{array} \right) 
    &= \left( \begin{array}{c} 
    \frac{1}{2} \left[ 1 - \pi^{-1} \arctan\left(y_{\text{P}}, x_{\text{P}} \right) \right] w_{\text{P}} \\ 
    \left[ m^{-1} \sqrt{x_{\text{P}}^2 + y_{\text{P}}^2} \right] h_{\text{P}} 
    \end{array} \right) \label{eq:lidarp}
\end{align}
where $\left(h_{\text{P}}, w_{\text{P}}\right)$ represents the size of LiDAR polar BEV, and $m$ denotes the maximum perception range. To ensure robustness, we represent each cell $\left(u_{\text{P}}, v_{\text{P}}\right)$ by its point density, which is calculated from the number of points within it. Similarly, the radar point cloud $\mathcal{R}$ can be transformed into radar polar BEV $\mathcal{B}_\text{R}$ with a shape of $\left(h_{\text{R}}, w_{\text{R}}\right)$.

Similarly, BEVPlace++{~\cite{luo2024bevplace++}} uses point density to construct Cartesian BEV representations. However, as viewpoint rotation causes nonlinear transformations in Cartesian BEV, BEVPlace++ necessitates the use of computationally expensive group convolutions to extract rotation-equivariant features. Thus, we instead select polar BEV. Its inherent property of translating viewpoint rotations into column shifts provides the data foundation for effective and efficient rotation-invariant place recognition (see Sec.~\ref{sec:rotation}).


\textbf{LiDAR/radar encoders.} To ensure the lightweight nature, we choose the ResNet18~\cite{he2016deep} as the backbone. To accommodate the resolution discrepancy between LiDAR and radar polar BEVs, we implement asymmetric ResNet encoders featuring a differing number of ResBlocks. By feeding the polar BEVs into the encoders, we can obtain spatially aligned and shape-matched feature maps. 

Our resulting feature maps also exhibit rotation-equivariant properties. Although spatial downsampling breaks our encoders' strict shift equivariance, making a mathematically explicit demonstration of rotation-equivariance impossible, the nature of polar BEV allows us to overcome this easily through the metric learning process. By training with viewpoint-diverse triplets, the network regains robustness to input translation, thereby achieving rotation-equivariant feature extraction.


\textbf{Descriptor extraction.} We employ NetVLAD{~\cite{arandjelovic2016netvlad}} to aggregate descriptors from both branches. Our goal is to leverage NetVLAD's permutation invariance{~\cite{uy2018pointnetvlad}} to extract rotation-invariant descriptors from rotation-equivariant feature maps, as demonstrated in{~\cite{ma2022overlaptransformer}}. This completes the final step for rotation-invariant place recognition. By concatenating the descriptors from both branches, place recognition with both accuracy and rotation-invariance can be achieved.

\subsection{Multimodal Feature Fusion}
\label{sec:feature_fusion}
\begin{figure}[t]
  \centering
  \includegraphics[width=1\linewidth]{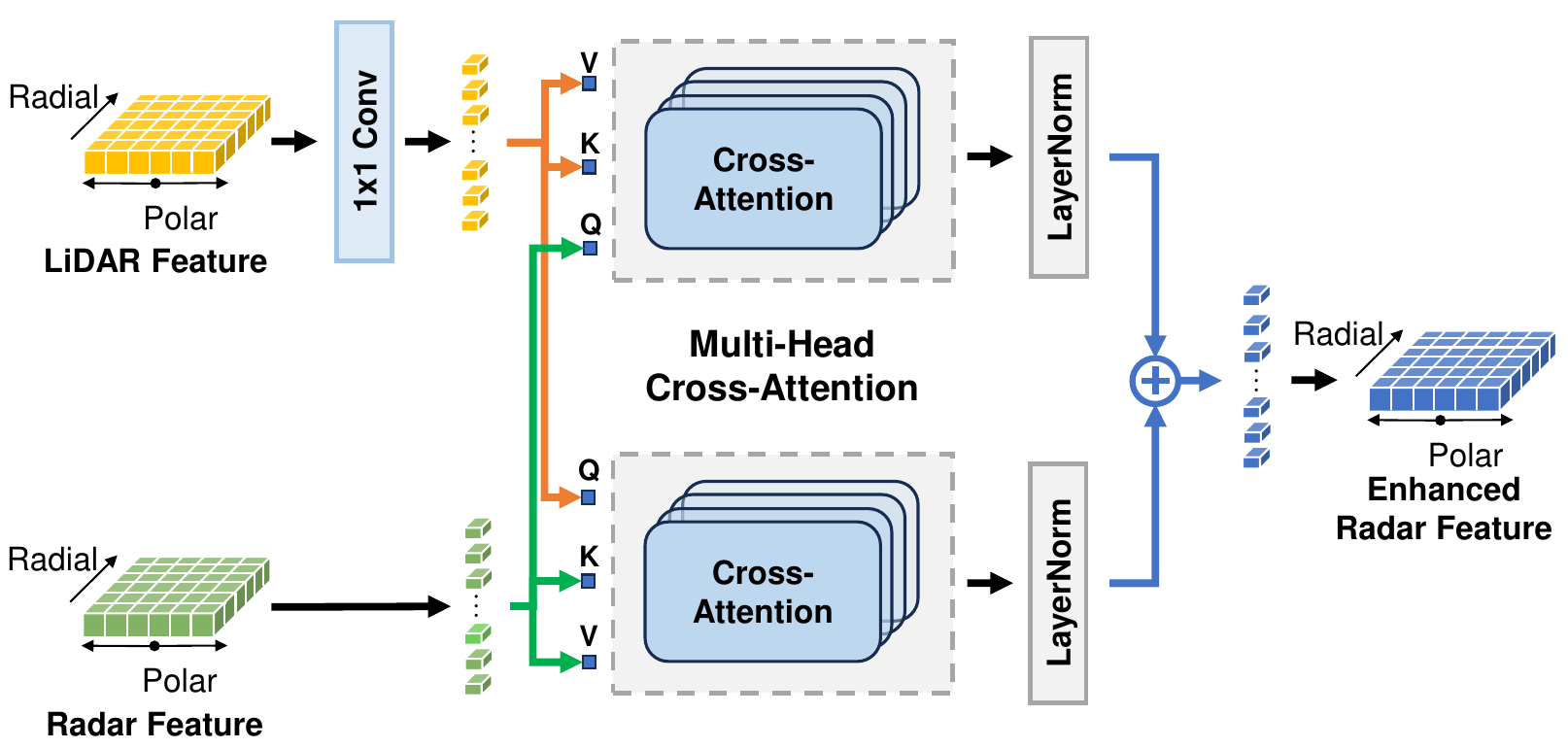}
  \caption{The pipeline of the polar cross-attention module. It consists of two cross-attention branches to facilitate feature interaction of the feature map, thus exploiting cross-modality correlations.}
  \label{fig:pattn}
  \vspace{-0.8cm}
\end{figure}

\textbf{Polar cross-attention module.} In MPR, directly fusing less discriminative modalities may impair performance (see Sec.~\mbox{\ref{sec:ablation}}). Thus, a key challenge in LiDAR-radar fusion, especially with single-chip radar, is mitigating data sparsity and noise to uncover discriminative radar features. To extract discriminative radar features, we develop a polar cross-attention module. It captures cross-modality correlations while enhancing the discriminability of radar features through attentive fusion, as detailed in~\figref{fig:pattn}.

To facilitate cross-attention, we first align the dimensions of the feature maps. We first compress LiDAR feature channels through convolution. This ensures the LiDAR feature matches the radar feature in dimensions. Next, we flatten both features into sequences $\mathcal{H}_\text{P} \in \mathbb{R}^{B \times HW \times C_{\text{R}}}$ and $\mathcal{H}_\text{R} \in \mathbb{R}^{B \times HW \times C_{\text{R}}}$, which are then processed by cross-attention for multimodal fusion. 

We design the cross-attention mechanism based on the multi-head self-attention (MHSA) module~\cite{vaswani2017attention}. The self-attention mechanism can be formulated as:
\begin{align}
    \mathrm{Attention}(\mathcal{Q},\mathcal{K},\mathcal{V})=\mathrm{softmax}\left(\frac{\mathcal{Q}\mathcal{K}^{\mathrm{T}}}{\sqrt{d_{\text{k}}}}\right)\mathcal{V}
\end{align}
where $\mathcal{Q},\mathcal{K},\mathcal{V}$ represent the queries, keys and values respectively, and $d_{\text{k}}$ represents the dimension of keys. We denote the enhanced radar feature extracted by the multi-head cross-attention as $\mathcal{A}$, and the cross-attention mechanism can be formulated as:
\begin{align}\label{eq4}
\begin{split}
    \bar{\mathcal{A}}&= \mathrm{LN}\left(\mathrm{Attention}\left(\mathcal{Q}_\text{R}, \mathcal{K}_\text{P}, \mathcal{V}_\text{P}\right)\right) \\
    &+ \mathrm{LN}\left(\mathrm{Attention}\left(\mathcal{Q}_\text{P}, \mathcal{K}_\text{R}, \mathcal{V}_\text{R}\right)\right) \\
\end{split}
\end{align}
where $\bar{\mathcal{A}}$ denotes the feature split of $\mathcal{A}$, $\mathrm{LN}(.)$ refers to layer normalization, $\mathcal{Q}_\text{R}$, $\mathcal{K}_\text{R}$, $\mathcal{V}_\text{R}$ are substantially the query, key, and value splits obtained by linearly transforming the splits of $\mathcal{H}_\text{R}$, and $\mathcal{Q}_\text{P}$, $\mathcal{K}_\text{P}$, $\mathcal{V}_\text{P}$ are those derived by the splits of $\mathcal{H}_\text{P}$.

Ultimately, the cross-attention enhanced radar feature is concatenated with LiDAR feature to generate the descriptor of the fusion branch. The proposed cross-attention mechanism extracts more discriminative feature representations by capturing cross-modal correlations.



\textbf{Structure-aware distillation.} 
The cross-modality feature fusion facilitates the extraction of discriminative multimodal place descriptions. However, while deep attentive fusion extracts discriminative features, we found it inadequate for fully exploiting radar's robustness (see Sec.~\ref{sec:ablation}). To address this, we propose the radar-exclusive distillation branch. Furthermore, to prevent less discriminative radar descriptors from negatively impacting recognition performance, we enhance their discriminability via cross-modality distillation.

We aim to improve the performance of the distillation branch through self-distillation. The structure-aware loss~\cite{knights2022incloud} (formally defined in~\ref{sec:network_training}) transfers geometric knowledge from the fusion branch by enforcing angular constraints on descriptor triplets. This geometric constraint ensures the descriptor triplets maintain their distributions while assimilating structural knowledge from the fusion branch. This enables the distillation branch to learn the discriminability of multimodal fusion features while exploiting the radar-specific robustness (see Sec.~\mbox{\ref{sec:ablation}}).

\subsection{Network Training}
\label{sec:network_training}
During the training process of our LRFusionPR, we jointly optimize the model using two different losses.  

\textbf{Triplet margin loss.} Following previous works~\cite{uy2018pointnetvlad, komorowski2021egonn}, we employ triplet loss to constrain descriptor distributions in high-dimensional space. For each query multimodal descriptor $\mathcal{D}_\text{q}$, we select one positive sample $\mathcal{D}_\text{pos}$ and $k$ negative samples $\mathcal{D}_\text{neg}$ to form a training triplet. We use Euclidean distance to distinguish positive and negative samples. Following AutoPlace~\cite{cait2022autoplace}, we define a positive region within 9\,m and a negative region beyond 18\,m. To enhance training efficiency, we apply lazy triplet loss~\cite{uy2018pointnetvlad} for mini-batch hard negative mining. The loss function is given by:
\begin{align}
    \mathcal{L}_{\text{LT}}=\left[\alpha+d(\mathcal{D}_\text{q},\mathcal{D}_{\text{pos}})-\mathop{\max}_{g}(d(\mathcal{D}_\text{q},\mathcal{D}_{\text{neg}}^{g}))\right]_+
\end{align}
where $\left [ \cdots \right ]_+$ denotes the hinge loss, $d(\cdot)$ means the Euclidean distance between a pair of descriptors, and $\alpha$ is the predefined margin.

\textbf{Structure-aware loss.} We use structure-aware loss~\cite{knights2022incloud} to supervise the knowledge transfer process from the fusion branch to the distillation branch. Given descriptor triplets $\mathcal{T}_{\text{F}}=\{\mathcal{D}_\text{F,q}, \mathcal{D}_\text{F,pos}, \mathcal{D}_\text{F,neg}\}$ and $\mathcal{T}_{\text{A}}=\{\mathcal{D}_\text{A,q}, \mathcal{D}_\text{A,pos}, \mathcal{D}_\text{A,neg}\}$ from the fusion and distillation branches, respectively. The structure-aware loss is defined as:
\begin{align}
    \mathcal{L}_{\text{SA}}=\mathrm{max}\left\{h\left(\phi_{\text{F}}, \phi_{\text{A}}\right)-\beta, 0\right\}
\end{align}
where $\beta$ is a predefined margin, $h$ is the Huber loss, expressed as follows:
\begin{align}
h(x,y)=\left\{
\begin{aligned}
&\frac{1}{2}(x-y)^{2},\ |x-y|\leq1\\
&|x-y|-\frac{1}{2},\ |x-y|>1
\end{aligned}
\right.\end{align}
and $\phi_{\text{F}}$ and $\phi_{\text{A}}$ is calculated by:
\begin{align}
    \phi_{\text{F}}=\langle\frac{\mathcal{D}_\text{F,pos} - \mathcal{D}_\text{F,q}}{||\mathcal{D}_\text{F,pos} - \mathcal{D}_\text{F,q}||_{2}}, \frac{\mathcal{D}_\text{F,neg} - \mathcal{D}_\text{F,q}}{||\mathcal{D}_\text{F,neg} - \mathcal{D}_\text{F,q}||_{2}}\rangle
\end{align}
\begin{align}
    \phi_{\text{A}}=\langle\frac{\mathcal{D}_\text{A,pos} - \mathcal{D}_\text{A,q}}{||\mathcal{D}_\text{A,pos} - \mathcal{D}_\text{A,q}||_{2}}, \frac{\mathcal{D}_\text{A,neg} - \mathcal{D}_\text{A,q}}{||\mathcal{D}_\text{A,neg} - \mathcal{D}_\text{A,q}||_{2}}\rangle
\end{align}
where $\langle,\rangle$ represents the inner product operation.

The total loss used for training LRFusionPR is obtained through a weighted combination of the lazy triplet loss and structure-aware loss, which is defined as:
\begin{align}
    \mathcal{L}_{\text{sum}}=\mathcal{L}_{\text{LT}}+\lambda\mathcal{L}_{\text{SA}}
\end{align}
where $\lambda$ is the weighting parameter.

\section{Experiments}
\label{sec:experiments}

\subsection{Datasets and Experimental Setup}
\label{sec:datasets}
To perform a comprehensive validation, we conduct experiments on four autonomous driving datasets: nuScenes{~\cite{caesar2020nuscenes}}, MulRan{~\cite{kim2020mulran}}, Oxford Radar{~\cite{barnes2020oxford}}, and Boreas{~\cite{burnett2023boreas}}.

\textbf{nuScenes.} The dataset features rainy weather and multiple revisits, with 32-beam LiDAR and single-chip radar data. Following AutoPlace~\cite{cait2022autoplace}, we train and evaluate the model on the BostonSeaport (BS) split. In addition, we validate the generalization ability on unseen splits, namely SingaporeOneNorth (SON) and SingaporeQueenstown (SQ). 

\textbf{MulRan.} The dataset emphasizes long revisit intervals and significant viewpoint transformations, with 64-beam LiDAR and scanning radar data. We perform evaluation on the Sejong, Riverside and DCC splits. We organize the Sejong split according to~\cite{komorowski2021egonn}, and the DCC split following~\cite{vidanapathirana2023spectral}. For the Riverside split, we use session 01 as database and session 02 as query. Due to the scanning radar's lower sampling frequency, we discard LiDAR data not time-synchronized with the radar (time difference over 0.05 s) to ensure multimodal data consistency.

\textbf{Oxford Radar.} Collected along a specific route in Oxford, the dataset includes 32-beam LiDAR and scanning radar data with temporal variations. We select the clear session ``2019-01-11-13-24-51" as the \textit{database set}, and the rainy session ``2019-01-16-14-15-33" as the \textit{test query set}. We choose these sessions (OxfordRadar-Rain) to evaluate our method's cross-dataset generalization capability in rainy conditions.

\textbf{Boreas.} The dataset provides real-world adverse weather data, including rain and snow, collected with a 128-beam LiDAR and scanning radar. We selected the clear-weather session ``2020-12-18-13-44" as our database. The ``2021-01-26-11-22" (Bor-Snow) and ``2021-04-29-15-55" (Bor-Rain) sessions, collected during snowy and rainy weather respectively, served as query sessions. We utilize this dataset to demonstrate our method's cross-dataset generalization capability under real-world adverse weather. More details of our training and evaluation splits are shown in~{\tabref{tab:dataset}}.

\subsection{Implementation Details}
\label{sec:expe_setup}
We configure the size of the LiDAR polar BEV as $h_{\text{P}}=200$ and $w_{\text{P}}=900$, and the radar polar BEV as $h_{\text{R}}=50$ and $w_{\text{R}}=225$, with 80\,m perception range $m$. For the NuScenes radar data, we stack seven nearest frames before projecting them into polar BEVs following~\cite{cait2022autoplace}. We apply height-based ground filtering to process LiDAR data. For the polar cross-attention module, we set the feature embedding dimension $d_{\text{model}}=128$, the feed-forward dimension $d_{\text{ffn}}=1024$, and the number of heads $n_{\text{heads}}=8$. For the NetVLAD module, we set the number of clusters $d_{\text{clusters}}=64$, and the dimension of the descriptor $N_{\text{D}}=256$. The final descriptor then has a size of $1\times512$. We set the margin of the Lazy Triplet Loss to $\alpha=0.5$, the margin of the structure-aware loss to $\beta=0.0002$, the weighting parameter $\lambda=5$, and the number of negative samples to $k=10$. We optimize the model using the Adam optimizer with an initial learning rate of $5 \times 10^{-5}$. The learning rate is halved every 5 epochs. We apply random translation and rotation for data augmentation. Given the more discriminative characteristics of scanning radar, we further adopt the distillation weight decay strategy from~{\cite{knights2022incloud}} when training on the MulRan dataset~{\cite{kim2020mulran}}. All of our experiments are conducted on a system with an Intel i7-14700KF CPU and an NVIDIA GeForce RTX 4060Ti GPU.


\begin{table}[t]
  \centering
  \begin{center}
  	\vspace{0.0cm}
  	\setlength{\tabcolsep}{9pt}
  	\renewcommand\arraystretch{1.0}
        \setlength{\abovecaptionskip}{0.15cm}
    \caption{Dataset statics and organizations}
    \resizebox{\columnwidth}{!}{
    \footnotesize{
        \begin{tabular}{c|c|c|c}
          \toprule
          Session & \makecell{Train \\ Db/Query} & \makecell{Test \\ Db/Query} & Step \\ \hline
          nuScenes-BS~\cite{caesar2020nuscenes} & 6312/7075 & 6312/3696 & 0.2\,m \\
          nuScenes-SON~\cite{caesar2020nuscenes} & 0/0 & 4007/1184 & 0.2\,m \\
          nuScenes-SQ~\cite{caesar2020nuscenes} & 0/0 & 418/851 & 0.2\,m \\
          MulRan-Sejong~\cite{kim2020mulran} & 14327$^{*}$ & 1505/1380 & 0.2\,m \\ 
          MulRan-Riverside~\cite{kim2020mulran} & 0/0 & 1558/1724 & 1.0\,m \\ 
          MulRan-DCC~\cite{kim2020mulran} & 0/0 & 204/129 & 10.0\,m \\
          OxfordRadar-Rain~\cite{barnes2020oxford} & 0/0 & 6361/5657 & 0.2\,m \\
          Bor-Rain~\cite{burnett2023boreas} & 0/0 & 3386/3350 & 0.2\,m \\
          Bor-Snow~\cite{burnett2023boreas} & 0/0 & 3386/4327 & 0.2\,m \\
        \bottomrule
        \multicolumn{4}{p{0.9\linewidth}}{$^{*}$The Sejong training split does not separate database and query sets.}\\
        \end{tabular}
        }
    }
    \label{tab:dataset}
    \end{center}
    \vspace{-0.9cm}
\end{table}

\begin{table*}[ht]
  \centering
  \begin{center}
  	\vspace{0.0cm}
  	\setlength{\tabcolsep}{3.9pt}
  	\renewcommand\arraystretch{0.95}
        \setlength{\abovecaptionskip}{0.15cm}
    \caption{Comparison of place recognition performance on the nuScenes dataset}
    \resizebox{\textwidth}{!}{
    \footnotesize{
        \begin{tabular}{cccccccccccccc}
          \toprule
          \multirow{2}{*}{Methods} & \multirow{2}{*}{Modality} &\multicolumn{4}{c}{BS split}&\multicolumn{4}{c}{SON split}&\multicolumn{4}{c}{SQ split} \\ \cline{3-14}
          ~ & ~ & AR@1 & AR@5 & AR@10 & max $F_{1}$ & AR@1 & AR@5 & AR@10  & max $F_{1}$ & AR@1 & AR@5 & AR@10  & max $F_{1}$ \\ \hline
          AutoPlace~\cite{cait2022autoplace} & R & 78.59 & 82.92 & 83.68 & 0.9453 & 78.36 & 86.90 & 90.45 & 0.9837 & 64.12 & 66.12 & 68.00 & 0.9411 \\
          MinkLoc3Dv2~\cite{komorowski2022improving} & L & 91.10 & 96.65 & 97.81 & 0.9557 & 97.21 & 99.32 & 99.41 & \underline{0.9880} & 91.07 & 98.71 & 99.18 & \underline{0.9565} \\
          EgoNN~\cite{komorowski2021egonn} & L & 89.80 & 96.43 & 97.59 & 0.9557 & 96.45 & 99.16 & 99.58 & 0.9841 & 85.90 & 96.83 & 98.00 & 0.9265 \\
          BEVPlace++~\cite{luo2024bevplace++} & L & \underline{97.27} & \underline{99.49} & \textbf{99.84} & 0.9370 & \underline{99.58} & \underline{100.00} & \underline{100.00} & 0.9686 & \underline{96.71} & \underline{99.29} & \underline{99.53} & 0.9267 \\
          MinkLoc++~\cite{komorowski2021minkloc++} & L+C & 82.01 & 91.29 & 93.64 & 0.9072 & 89.95 & 95.44 & 96.45 & 0.9605 & 64.75 & 81.79 & 87.07 & 0.8167 \\
          LCPR~\cite{zhou2023lcpr} & L+C & 83.01 & 90.83 & 93.61 & 0.9099 & 89.10 & 96.88 & 97.89 & 0.9424 & 61.34 & 75.56 & 80.14 & 0.8112 \\
          CRPlace~\cite{fu2024crplace} & C+R & 91.2 & 92.6 & 93.3 & \underline{0.96} & - & - & - & - & - & - & - & - \\
          \rowcolor{gray!25}
          LRFusionPR (ours) & L+R & \textbf{97.75} & \textbf{99.51} & \underline{99.73} & \textbf{0.9892} & \textbf{99.92} & \textbf{100.00} & \textbf{100.00} & \textbf{0.9996} & \textbf{97.65} & \textbf{99.65}  & \textbf{99.76} & \textbf{0.9899} \\
          \bottomrule
        \multicolumn{14}{p{0.9\linewidth}}{C: Camera, L: LiDAR, R: Single-Chip Radar. The best and secondary results are highlighted in \textbf{bold black} and \underline{underline} respectively.}\\
        \end{tabular}
        }
    }
    \label{tab:nuscenes}
    \end{center}
    \vspace{-0.6cm}
\end{table*}

\begin{table*}[ht]
  \centering
  \begin{center}
  	\vspace{0.15cm}
  	\setlength{\tabcolsep}{3.9pt}
  	\renewcommand\arraystretch{0.95}
        \setlength{\abovecaptionskip}{0.15cm}
    \caption{Comparison of place recognition performance on the MulRan dataset}
    \resizebox{\textwidth}{!}{
    \footnotesize{
        \begin{tabular}{cccccccccccccc}
          \toprule
          \multirow{2}{*}{Methods} & \multirow{2}{*}{Modality} &\multicolumn{4}{c}{Sejong}&\multicolumn{4}{c}{Riverside}&\multicolumn{4}{c}{DCC} \\ \cline{3-14}
         ~&~ & AR@1 & AR@5 & AR@10 & max $F_{1}$ & AR@1 & AR@5 & AR@10  & max $F_{1}$ & AR@1 & AR@5 & AR@10  & max $F_{1}$ \\ \hline
          KidnappedRadar~\cite{suaftescu2020kidnapped} & R & 70.14 & 81.88 & 87.32 & 0.8699 & 72.80 & 89.79 & 94.20 & 0.8446 & 55.04 & 70.54 & 78.29 & 0.7722 \\
          RadVLAD~\cite{gadd2024open} & R & 94.06 & 97.75 & 98.77 & 0.9784 & 83.99 & 92.69 & 94.84 & 0.9380 & 58.91 & 75.97 & 79.07 & 0.8553 \\
          FFT-RadVLAD~\cite{gadd2024open} & R & 88.41 & 93.77 & 95.36 & 0.9462 & 79.29 & 89.33 & 92.05 & 0.9112 & 49.61 & 72.87 & 77.52 & 0.7973 \\
          MinkLoc3Dv2~\cite{komorowski2022improving} & L & 88.70 & 97.46 & 98.41 & 0.9414 & 82.08 & 91.76 & 94.43 & 0.9019 & 66.67 & 81.40 & 84.50 & 0.8177 \\
          EgoNN~\cite{komorowski2021egonn} & L & 94.42 & 98.70 & 99.42 & 0.9715 & 79.00 & 86.48 & 89.56 & 0.9177 & 67.44 & 79.07 & 82.95 & 0.9050 \\
          EgoNN$^{\dag}$~\cite{komorowski2021egonn} & L & \underline{97.32} & \underline{99.71} & \underline{99.86} & \underline{0.9871} & 81.26 & 86.77 & 89.39 & \underline{0.9425} & 72.09 & 82.17 & 82.95 & 0.9045 \\
          BEVPlace++~\cite{luo2024bevplace++} & L & 92.75 & 97.03 & 97.39 & 0.9544 & \underline{91.30} & \underline{95.42} & \underline{96.40} & 0.9258 & \underline{73.64} & \underline{85.27} & \underline{89.15} & \underline{0.9118} \\
          \rowcolor{gray!25}
          LRFusionPR (ours) & L+R & \textbf{98.91} & \textbf{99.93} & \textbf{100.00} & \textbf{0.9945} & \textbf{95.19} & \textbf{98.20} & \textbf{99.01} & \textbf{0.9753} & \textbf{75.19} & \textbf{86.82} & \textbf{90.70} & \textbf{0.9406} \\
          \bottomrule
        \multicolumn{14}{p{0.9\linewidth}}{C: Camera, L: LiDAR, R: Scanning Radar. The best and secondary results are highlighted in \textbf{bold black} and \underline{underline} respectively.}\\
        \end{tabular}
        }
    }
    \label{tab:mulran}
    \end{center}
    \vspace{-0.6cm}
\end{table*}

\begin{table*}[ht]
  \centering
  \begin{center}
  	\vspace{0.15cm}
  	\setlength{\tabcolsep}{1.2pt}
  	\renewcommand\arraystretch{0.95}
        \setlength{\abovecaptionskip}{0.15cm}
    \caption{Place recognition performance under adverse weather conditions}
    \resizebox{\textwidth}{!}{
    \footnotesize{
        \begin{tabular}{cc|ccc|ccc|ccc|ccc|ccc}
          \toprule
          \multirow{3}{*}{Methods} & \multirow{3}{*}{Modality} & \multicolumn{6}{c|}{Rain} & \multicolumn{6}{c|}{Simulated Fog} & \multicolumn{3}{c}{Snow} \\ \cline{3-17}
          ~ & ~ & \multicolumn{3}{c|}{OxfordRadar-Rain} & \multicolumn{3}{c|}{Bor-Rain} & \multicolumn{3}{c|}{SQ-Fog} & \multicolumn{3}{c|}{Sejong-Fog} & \multicolumn{3}{c}{Bor-Snow} \\ \cline{3-17}
          ~ & ~ & AR@1 & AR@5 & AR@10 & AR@1 & AR@5 & AR@10 & AR@1 & AR@5 & AR@10 & AR@1 & AR@5 & AR@10 & AR@1 & AR@5 & AR@10 \\ \hline
          AutoPlace~\cite{cait2022autoplace} & CR & - & - & - & - & - & - & 64.12 & 66.12 & 68.00 & - & - & - & - & - & - \\
          KidnappedRadar~\cite{suaftescu2020kidnapped} & SR & 84.39 & 90.63 & 92.22 & 55.37 & 72.00 & 78.48 & - & - & - & 70.14 & 81.88 & 87.32 & 59.42 & 71.11 & 74.81 \\
          RadVLAD~{\cite{gadd2024open}} & SR & 86.69 & 92.87 & 94.83 & 96.20 & 97.89 & 98.47 & - & - & - & \textbf{94.06} & \underline{97.75} & \underline{98.77} & \underline{95.71} & 98.41 & 99.35 \\
          FFT-RadVLAD~\cite{gadd2024open} & SR & 93.94 & \underline{97.99} & 98.31 & 56.25 & 71.29 & 78.73 & - & - & - & 88.41 & 93.77 & 95.36 & 94.59 & \underline{98.82} & \underline{99.53} \\
          MinkLoc3Dv2~\cite{komorowski2022improving} & L & 92.68 & 97.35 & 98.32 & 94.36 & 98.24 & 99.28 & \underline{75.91} & \underline{91.42} & \underline{94.24} & 65.72 & 81.67 & 87.17 & 19.51 & 32.77 & 38.66 \\
          EgoNN~\cite{komorowski2021egonn} & L & 87.98 & 94.87 & 96.71 & 98.81 & 99.67 & 99.82 & 35.84 & 50.18 & 57.34 & 40.07 & 55.14 & 61.67 & 71.13 & 84.28 & 87.89 \\
          BEVPlace++~\cite{luo2024bevplace++} & L & \underline{94.36} & 97.51 & \underline{98.57} & \underline{99.46} & \underline{99.91} & \underline{99.94} & 67.57 & 76.73 & 81.08 & 56.16 & 70.94 & 76.74 & 88.42 & 94.34 & 96.07 \\
          MinkLoc++~\cite{komorowski2021minkloc++} & L+C & 93.07 & 97.19 & 97.88 & 85.67 & 96.45 & 97.97 & 23.15 & 46.89 & 58.28 & - & - & - & 3.51 & 8.32 & 14.01 \\
          LCPR~\cite{zhou2023lcpr} & L+C & - & - & - & - & - & - & 13.98 & 27.03 & 37.49 & - & - & - & - & - & - \\
          \rowcolor{gray!25}
          LRFusionPR (ours) & L+R & \textbf{98.44} & \textbf{99.29} & \textbf{99.42} & \textbf{100.00} & \textbf{100.00} & \textbf{100.00} & \textbf{91.07} & \textbf{96.47} & \textbf{97.53} & \underline{93.48} & \textbf{98.41} & \textbf{99.42} & \textbf{99.38} & \textbf{99.72} & \textbf{99.88} \\
          \bottomrule
        \multicolumn{17}{p{0.9\linewidth}}{C: Camera, L: LiDAR, CR: Single-Chip Radar, SR: Scanning Radar, R: Compatible with both Single-Chip and Scanning Radar.}\\
        \multicolumn{17}{p{0.9\linewidth}}{Missing entries in the table indicate that the session could not meet the input data requirements for the corresponding method.}\\
        \end{tabular}
        }
    }
    \label{tab:robust}
    \end{center}
    \vspace{-0.9cm}
\end{table*}

\subsection{Evaluation for Place Recognition}
\label{sec:eval_res}
\textbf{nuScenes results.} Following~\cite{cait2022autoplace, zhou2023lcpr}, we report AR@1, AR@5, AR@10, and max $F_{1}$ score as metrics. We compare LRFusionPR to SOTA baselines, including unimodal (MinkLoc3Dv2~{\cite{komorowski2022improving}}, EgoNN~{\cite{komorowski2021egonn}}, BEVPlace++~{\cite{luo2024bevplace++}}, AutoPlace~{\cite{cait2022autoplace}}, RadVLAD~{\cite{gadd2024open}}, FFT-RadVLAD~{\cite{gadd2024open}}) and multimodal ones (MinkLoc++~\cite{komorowski2021minkloc++}, LCPR~\cite{zhou2023lcpr}, CRPlace~\cite{fu2024crplace}). We train and evaluate baselines using their official implementations. To ensure a fair comparison, we extend the maximum perception range of BEVPlace++~\cite{luo2024bevplace++} from 40\,m to 80\,m to match other baselines. For CRPlace~\cite{fu2024crplace}, we report the results from their paper due to the lack of open-source code.

As shown in Table~{\ref{tab:nuscenes}}, our method demonstrates SOTA performance across various metrics and sessions. By leveraging the complementary strengths of LiDAR and radar, our approach outperforms the baselines. In addition, we observe that LiDAR-camera fusion shows little improvement over LPR baselines. In the AutoPlace benchmark, where the \textit{database set} contains only clear weather data and the \textit{query set} is collected in rain, we hypothesize this is due to overfitting on training images.

\textbf{MulRan results.} Since MulRan~{\cite{kim2020mulran}} does not include camera data, we use KidnappedRadar~{\cite{suaftescu2020kidnapped}}, RadVLAD~{\cite{gadd2024open}}, FFT-RadVLAD~{\cite{gadd2024open}}, MinkLoc3Dv2~{\cite{komorowski2022improving}}, EgoNN~{\cite{komorowski2021egonn}}, and BEVPlace++~{\cite{luo2024bevplace++}} as the baselines. The place recognition evaluation metrics follow the experiments on the nuScenes dataset. 
For a fair comparison, we train BEVPlace++, MinkLoc3Dv2 and EgoNN using~\cite{komorowski2021egonn}'s data partition (35.9k LiDAR training scans). We also evaluate the open-source EgoNN weights, denoted as $\text{EgoNN}^{\dag}$, which is derived from joint training over multiple datasets.

As shown in~{\tabref{tab:mulran}}, our method outperforms all baselines on Sejong split, and demonstrates superior generalization ability on Riverside and DCC splits. This further demonstrates that our method effectively generates discriminative and robust multimodal place descriptors.


\subsection{Robustness under Adverse Weather}
\label{sec:rob}
In this section, we perform experiments on place recognition under adverse weather. To comprehensively evaluate the robustness of our proposed method in adverse weather conditions, we conduct experiments under three distinct weather types: rain, fog, and snow. Specifically, we select OxfordRadar-Rain and Bor-Rain as two rain-affected scenarios, and Bor-Snow for snowy conditions (see Sec.~\mbox{\ref{sec:datasets}} for details). Due to the lack of open-source fog datasets suitable for place recognition, we simulate fog conditions using the fog model from~{\cite{bijelic2020seeing}} following MVDNet~{\cite{qian2021robust}}. We perform fog simulations on the Sejong and SQ sessions, which serve as our two fog test scenarios.

\begin{figure}[t]
\vspace{0.0cm}
  \centering
  \includegraphics[width=1\linewidth]{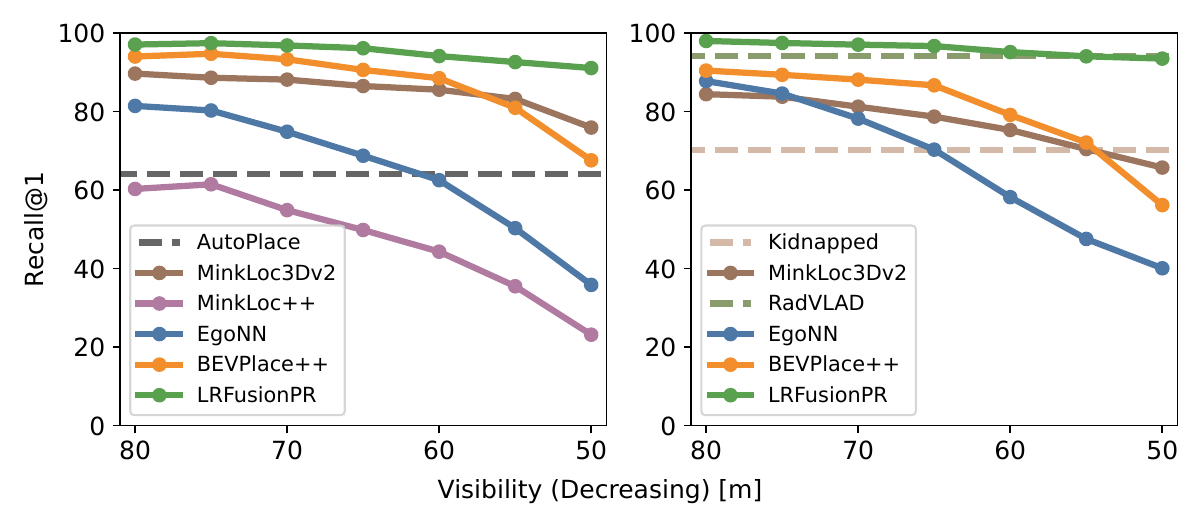}
  \caption{The variation of the Recall@1 as a function of visibility under fog on SQ (left) and Sejong (right) splits.}
  \label{fig:fog}
  \vspace{-0.75cm}
\end{figure}

\begin{figure}
  \centering
  \includegraphics[width=1\linewidth]{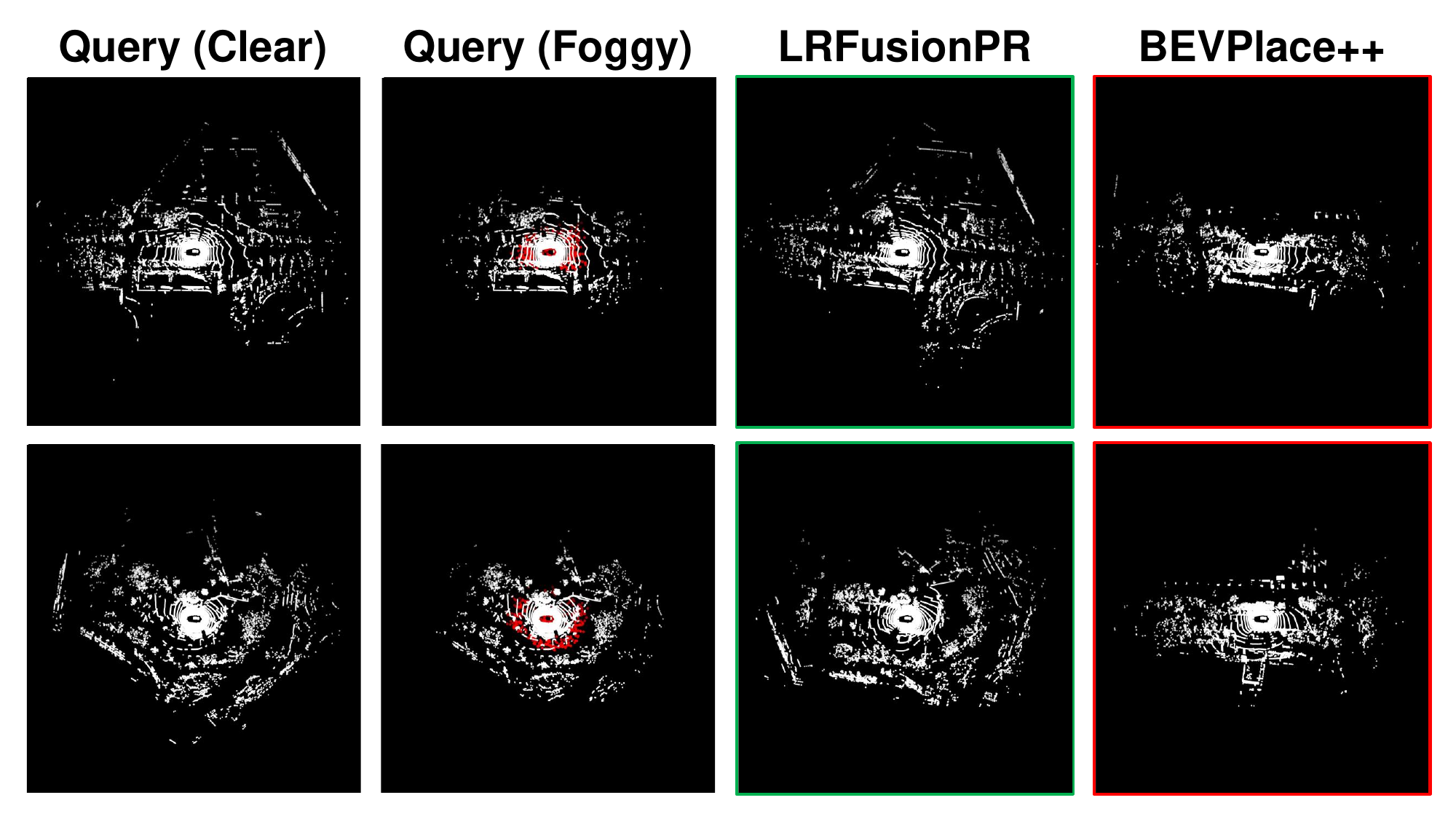}
  \caption{Visualization of place recognition in foggy weather. Each row in the figure corresponds to a different scene. The green borders indicate correct retrieval results, while the red borders represent incorrect retrievals. We use red points to highlight the noise introduced by fog.}
  \label{fig:fog_viz}
  \vspace{-0.75cm}
\end{figure}

{\tabref{tab:robust}} shows the recognition accuracy of our proposed method and the baselines across five scenarios representing different weather conditions. Additionally,~{\figref{fig:fog}} illustrates the variation of Recall@1 with visibility under simulated fog conditions. We find that LiDAR baselines suffer significant degradation in foggy and snowy conditions, where noise is prevalent and visibility is reduced. Although radar baselines demonstrate comparatively lower accuracy, they surpass LiDAR baselines in adverse weather. Notably, our method consistently outperforms both LiDAR-only, radar-only, and multimodal baselines across all adverse weather conditions, demonstrating the effectiveness of LiDAR-radar fusion for robust place recognition (see~{\figref{fig:fog_viz}} for visualization). Furthermore, since our training set did not include Boreas~{\cite{burnett2023boreas}} or Oxford Radar~{\cite{barnes2020oxford}} data, these results demonstrate the cross-dataset generalization capability of our LRFusionPR.

\begin{table*}[ht]
  \centering
  \begin{center}
  	\vspace{0.0cm}
  	\setlength{\tabcolsep}{4.9pt}
  	\renewcommand\arraystretch{0.95}
        \setlength{\abovecaptionskip}{0.15cm}
    \caption{Ablation study on the effect of proposed component}
    \resizebox{\textwidth}{!}{
    \footnotesize{
        \begin{tabular}{cccccccccccccccc}
          \toprule
          \multirow{2}{*}{RA} & \multirow{2}{*}{PC} & \multirow{2}{*}{DB} & \multirow{2}{*}{SA} & \multicolumn{4}{c}{Single-Chip Radar Dataset} & \multicolumn{4}{c}{Scanning Radar Dataset} & \multicolumn{4}{c}{Adverse Weather Dataset} \\ \cline{5-16}
          ~ & ~ & ~ & ~ & AR@1 & AR@5 & AR@10 & max $F_{1}$ & AR@1 & AR@5 & AR@10 & max $F_{1}$ & AR@1 & AR@5 & AR@10 & max $F_{1}$ \\ \hline
          - & - & - & - & 97.97 & 99.61 & 99.78 & 0.9906 & 89.74 & 94.38 & 96.01 & 0.9598 & 82.06 & 88.84 & 91.50 & 0.9098 \\
          \checkmark & - & - & - & 96.74 & 99.30 & 99.57 & 0.9851 & 87.33 & 93.08 & 94.52 & 0.9532 & 77.04 & 85.23 & 88.90 & 0.8827 \\
          \checkmark & \checkmark & - & - & 97.96 & 99.60 & 99.82 & 0.9902 & 89.81 & 95.02 & 95.89 & 0.9586 & 83.81 & 90.65 & 93.39 & 0.9248 \\
          \checkmark & \checkmark & \checkmark & - & 98.03 & 99.63 & 99.85 & 0.9908 & \textbf{91.94} & \textbf{96.54} & \textbf{97.38} & 0.9707 & 93.57 & 97.61 & 98.57 & 0.9694 \\
          \rowcolor{gray!25}
          \checkmark & \checkmark & \checkmark & \checkmark & \textbf{98.44} & \textbf{99.72} & \textbf{99.95} & \textbf{0.9929} & 91.93 & 96.06 & 97.28 & \textbf{0.9759} & \textbf{95.19} & \textbf{98.42} & \textbf{99.03} & \textbf{0.9772} \\
          \bottomrule
        \end{tabular}
        }
    }
    \label{tab:ablation_st}
    \end{center}
    \vspace{-0.6cm}
\end{table*}

\begin{table*}[ht]
  \centering
  \begin{center}
  	\vspace{0.15cm}
  	\setlength{\tabcolsep}{5.8pt}
  	\renewcommand\arraystretch{0.95}
        \setlength{\abovecaptionskip}{0.15cm}
    \caption{Ablation study on input modalities}
    \resizebox{\textwidth}{!}{
    \footnotesize{
        \begin{tabular}{c|cc|cc|cc|cc|cc|cc}
          \toprule
          \multirow{2}{*}{Method} & \multicolumn{2}{c|}{SQ} & \multicolumn{2}{c|}{Sejong} & \multicolumn{2}{c|}{Riverside} & \multicolumn{2}{c|}{OxfordRadar-Rain} & \multicolumn{2}{c|}{SQ-Fog} & \multicolumn{2}{c}{Bor-Snow} \\ \cline{2-13}
          ~ & AR@1 & AR@10 & AR@1 & AR@10 & AR@1 & AR@10 & AR@1 & AR@10 & AR@1 & AR@10 & AR@1 & AR@10 \\ \hline
          LRFusionPR-L & 95.77 & 99.65 & 97.46 & 99.64 & 87.99 & 95.30 & 96.52 & 99.13 & 64.98 & 82.61 & 88.63 & 95.68 \\
          LRFusionPR-R & 78.38 & 91.54 & 93.26 & 97.83 & 94.78 & 99.30 & 95.26 & 98.30 & 78.38 & 91.54 & 98.54 & 99.45 \\
          \rowcolor{gray!25}
          LRFusionPR & \textbf{97.65} & \textbf{99.76} & \textbf{98.91} & \textbf{100.00} & \textbf{95.19} & \textbf{99.01} & \textbf{98.44} & \textbf{99.42} & \textbf{91.07} & \textbf{97.53} & \textbf{99.38} & \textbf{99.88} \\
          \bottomrule
        \end{tabular}
        }
    }
    \label{tab:ablation_im}
    \end{center}
    \vspace{-0.8cm}
\end{table*}


\subsection{Ablation Study}
\label{sec:ablation}
In this section, we conduct ablation studies on the network architecture, input modalities, and the impact of multimodal training. We aim to demonstrate the rationale of the network design, and the contributions of each individual modalities for accurate and robust place recognition.

\textbf{Network architecture.} We analyze the impact of four key components: radar input (RA), polar cross-attention (PC), distillation branch (DB), and structure-aware loss (SA). For a comprehensive ablation, we construct three multi-split datasets: single-chip radar (from nuScenes), scanning radar (from MulRan and Oxford Radar), and adverse weather (integrating BS, OxfordRadar-Rain, SQ-Fog, and Sejong-Fog). We then compute the average metrics across all splits within each dataset. All different network architectures were trained from scratch. Our LiDAR-only baseline retains only the architecture associated with LiDAR input. As shown in~\tabref{tab:ablation_st} (row 1), adverse weather severely degrades its recognition accuracy. Direct introduction of radar input via a simple concatenation-convolution strategy failed to exploit discriminative radar features, leading to a loss of accuracy (row 2). Implementing polar cross-attention for multimodal fusion increases robustness without sacrificing accuracy (row 3). Further integration of the distillation branch significantly improves weather resilience  (row 4). Finally, structure-aware distillation boosts the discriminability of radar-exclusive descriptors, yielding additional robustness gains. This validates the rationale of our network architecture design.

\textbf{Input modalities.} To elucidate each modality's role in our architecture, we conduct an ablation study on input modalities. We compare three input configurations: LiDAR-only (LRFusionPR-L), radar-only (LRFusionPR-R), and the full LRFusionPR. As shown in~\mbox{\tabref{tab:ablation_im}}, while unimodal models can achieve comparable performance to the baselines, the superiority of LRFusionPR primarily stems from synergetic multimodal fusion. Even in clear weather with discriminative LiDAR data (SQ, Sejong), radar fusion yields further improvements. Meanwhile, in scenarios requiring long-range perception (Riverside), our approach effectively leverages the advantages of radar. In adverse weather (OxfordRadar-Rain, SQ-Fog, Bor-Snow), the LiDAR-only variant clearly degrades, especially in fog and snow. However, LRFusionPR fully exploits radar's robustness, achieving robust place recognition through modal complementarity.

\textbf{Impact of multimodal training.} Here we validate the enhancement of radar modality through multimodal training. We compare the distillation branch trained with multimodal data (LRFusionPR-R-M) with that trained solely on radar data (LRFusionPR-R-S). As shown in~\mbox{\tabref{tab:ablation_mm}}, LRFusionPR-R-M outperforms its single-modality trained counterpart LRFusionPR-R-S. This improvement stems from the knowledge learned via structure-aware distillation during multimodal training. The benefit is particularly evident on several single-chip radar sequences (BS, SON, SQ).


\subsection{Study on Viewpoint Rotation Invariance}
\label{sec:rotation}
\begin{figure}[t]
\vspace{0.4cm}
  \centering
  \includegraphics[width=1\linewidth]{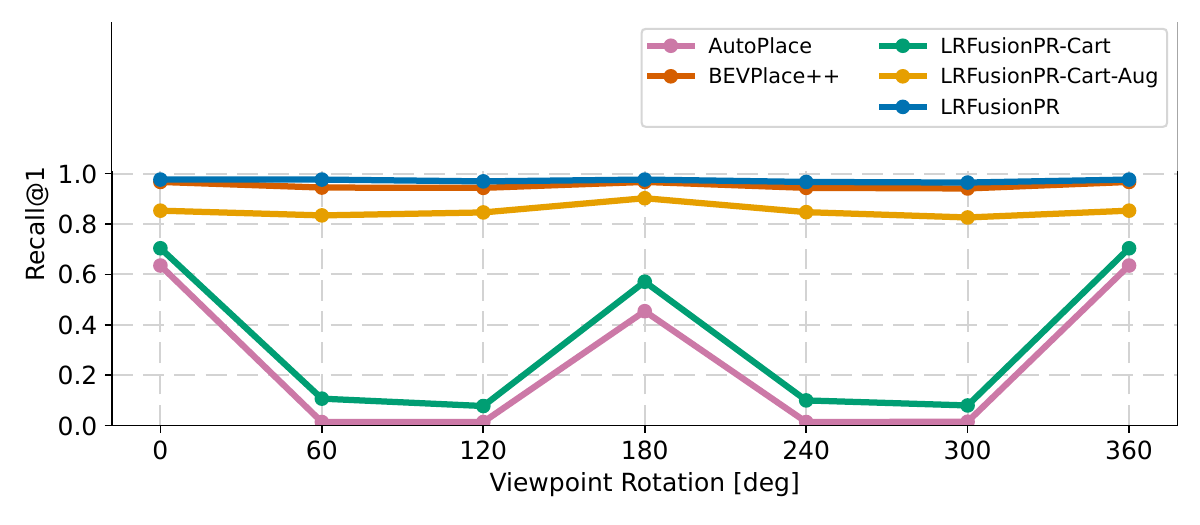}
  \caption{Comparative analysis of rotation invariance across baselines.}
  \label{fig:rotation}
  \vspace{-0.7cm}
\end{figure}

In this section, we evaluate the viewpoint rotation invariance property of our proposed LRFusionPR. Following~\cite{ma2022overlaptransformer, zhou2023lcpr}, we conduct experiments by rotating each query sample by $\left[60°, 120°, 180°, 240°, 300°\right]$.~\figref{fig:rotation} illustrates the rotation robustness of BEV-based baselines, LRFusionPR, and its variants on the SQ split. AutoPlace~\cite{cait2022autoplace} and LRFusionPR-Cart, both using Cartesian BEV as input, exhibit significant viewpoint sensitivity. LRFusionPR-Cart-Aug mitigates this issue through data augmentation, but does not achieve complete rotation invariance. BEVPlace++~\cite{luo2024bevplace++} achieves rotation invariance via the computationally expensive group convolution architecture combined with data augmentation. In contrast, leveraging the properties of polar BEV, our LRFusionPR achieves strong viewpoint rotation robustness with only simple data augmentation, eliminating the need for specially designed modules. This validates our approach of exploiting polar BEV's inherent characteristics for rotation-invariant place recognition.


\begin{table}[t]
  \centering
  \vspace{0.1cm}
  \setlength{\tabcolsep}{4.5pt}
  \renewcommand\arraystretch{0.95}
  \setlength{\abovecaptionskip}{0.15cm}
  \caption{Ablation study on multimodal training}
  \resizebox{\columnwidth}{!}{
  \footnotesize{
    \begin{tabular}{c|ccc|>{\columncolor{gray!25}}c >{\columncolor{gray!25}}c >{\columncolor{gray!25}}c}
        \toprule
        \multirow{2}{*}{Session} & \multicolumn{3}{c|}{LRFusionPR-R-S} & \multicolumn{3}{c}{\cellcolor{gray!25}LRFusionPR-R-M} \\ \cline{2-7}
        ~ & AR@1 & AR@5 & AR@10 & AR@1 & AR@5 & AR@10 \\ \hline
        BS & 50.60 & 63.77 & 69.21 & \textbf{61.44} & \textbf{74.43} & \textbf{78.68} \\
        SON & 56.76 & 70.78 & 75.34 & \textbf{61.15} & \textbf{76.52} & \textbf{83.36} \\
        SQ & 69.33 & 82.61 & 86.84 & \textbf{78.38} & \textbf{91.54} & \textbf{94.48} \\
        Sejong & 92.39 & 96.45 & 97.39 & \textbf{93.26} & \textbf{96.81} & \textbf{97.83} \\
        Riverside & 94.08 & 98.03 & 98.55 & \textbf{94.78} & \textbf{98.67} & \textbf{99.30} \\
        Bor-Snow & 98.15 & 98.78 & 99.12 & \textbf{98.54} & \textbf{99.19} & \textbf{99.45} \\
        \bottomrule
    \end{tabular}
    }
  }
  \label{tab:ablation_mm}
  \vspace{-0.6cm}
\end{table}

\begin{table}[t]
  \centering
  \setlength{\tabcolsep}{1.5pt}
  \renewcommand\arraystretch{0.95}
  \setlength{\abovecaptionskip}{0.15cm}
  \caption{Runtime analysis and comparison}
  \resizebox{\columnwidth}{!}{
  \footnotesize{
    \begin{tabular}{c|c|c|c|c}
        \toprule
        Method & Extraction [ms] & Retrieval [ms] & Total [ms] & GFLOPs \\ \hline
        AutoPlace~\cite{cait2022autoplace} & 21.08 & 4.22 & 25.30 & 20.77 \\
        MinkLoc3Dv2~\cite{komorowski2022improving} & 7.36 & \textbf{0.09} & 7.45 & - \\
        EgoNN~\cite{komorowski2021egonn} & 9.41 & \textbf{0.09} & 9.50 & - \\
        BEVPlace++~\cite{luo2024bevplace++} & 17.99 & 8.43 & 26.42 & 10.84 \\
        MinkLoc++~\cite{komorowski2021minkloc++} & 17.14 & 0.10 & 17.24 & - \\
        LCPR~\cite{zhou2023lcpr} & 8.38 & 0.12 & 8.50 & 57.85 \\
        \rowcolor{gray!25}
        LRFusionPR & \textbf{6.53} & 0.25 & \textbf{6.78} & \textbf{7.22} \\
        \bottomrule
    \multicolumn{5}{p{1.0\columnwidth}}{GFLOPS for sparse convolution-based baselines are not reported due to unmeasurable complexity.}\\
    \end{tabular}
    }
  }
  \label{tab:efficiency}
  \vspace{-0.7cm}
\end{table}

\subsection{Efficiency Analysis}
\label{sec:efficiency}
In this section, we provide an efficiency analysis of our LRFusionPR. We compare the runtime and computational complexity of LRFusionPR with the baselines. We report the average time of descriptor extraction and similarity retrieval on the BS split. 
As shown in~\tabref{tab:efficiency}, LRFusionPR only takes 6.78\,ms to perform a place recognition
, demonstrating its real-time capability. Additionally, our LRFusionPR achieves low memory consumption and computational complexity with 24.4\,M parameters and 7.22\,GFLOPs.

\section{Conclusion}
\label{sec:conclusion}
In this paper, we propose a LiDAR-radar fusion place recognition network applicable to both single-chip and scanning radar. We unify multi-sensor data into polar BEVs, eliminating domain discrepancies while enabling rotation-invariant place recognition. Furthermore, our proposed dual-branch network then leverages deep attentive fusion and cross-modality self-distillation to extract discriminative representations from noisy radar data. Validated across four datasets and diverse weather conditions, our LRFusionPR consistently outperforms all baselines, demonstrating multimodal complementarity for accurate and robust recognition.


\bibliographystyle{ieeetran}

\footnotesize{
\bibliography{glorified, new}}

\end{document}